\documentclass{article}

\usepackage{url}
\usepackage{cite}
\usepackage{xspace}
\usepackage{rotating}
\usepackage{multirow}
\usepackage{mathptmx}
\usepackage{graphicx}
\usepackage{subfigure}
\usepackage{amsmath}
\usepackage{amssymb}
\usepackage{times}
\usepackage{dsfont}
\usepackage{clrscode}
\usepackage{fancyvrb}
\usepackage[usenames,dvipsnames]{xcolor}
\usepackage{float}
\usepackage{algorithmic}
\usepackage{algorithm}
\usepackage{microtype}
\newfloat{algorithm}{t}{lop}

\usepackage{color}

\newfloat{algorithm}{t}{lop}

\newcommand{\vfkm}{vector field $k$-means\xspace}
\newcommand{\VFKM}{Vector field $k$-means\xspace}
\newcommand{\km}{$k$-means\xspace}

\newcommand{\hide}[1]{}
\newcommand{\claudio}[1]{}
\newcommand{\nivan}[1]{}
\newcommand{\carlos}[1]{}
\newcommand{\jim}[1]{}
\newcommand{\hidecomment}[1]{}

\newcommand{\ie}{i.e.}

\newcommand{\fitVectorField}{\textrm{fitVectorField}}
{\begin{list}{$\bullet$}{%
	\setlength{\labelsep}{2pt}\setlength{\leftmargin}{0pt}%
	\setlength{\labelwidth}{0pt}%
	\setlength{\listparindent}{0pt}}}
{\end{list}}

\graphicspath{{../figs_small/}}

\begin{document}

\markboth{N. Ferreira et al.}{Vector Field $k$-Means: Clustering Trajectories by Fitting Multiple Vector Fields}

\title{Vector Field $k$-Means: Clustering Trajectories by Fitting Multiple Vector Fields} 

\author{Nivan Ferreira \and Claudio Silva \and James T. Klosowski \and Carlos Scheidegger}




\maketitle



\begin{abstract} 
Scientists study trajectory data to understand trends in movement patterns, such as human mobility for traffic analysis and urban planning. There is a pressing need for scalable and efficient techniques for analyzing this data and discovering the underlying patterns. In this paper, we introduce a novel technique which we call vector-field $k$-means.

The central idea of our approach is to use vector fields to induce a similarity notion between trajectories. Other clustering algorithms seek a representative trajectory that best describes each cluster, much like $k$-means identifies a representative ``center'' for each cluster. Vector-field $k$-means, on the other hand, recognizes that in all but the simplest examples, no single trajectory adequately describes a cluster. Our approach is based on the premise that movement trends in trajectory data can be modeled as flows within multiple vector fields, and the \emph{vector field itself} is what defines each of the clusters. We also show how vector-field $k$-means connects techniques for scalar field design on meshes and $k$-means clustering.

We present an algorithm that finds a locally optimal clustering of trajectories into vector fields, and demonstrate how vector-field $k$-means can be used to mine patterns from trajectory data. We present experimental evidence of its effectiveness and efficiency using several datasets, including historical hurricane data, GPS tracks of people and vehicles, and anonymous call records from a large phone company. We compare our results to previous trajectory clustering techniques, and find that our algorithm performs faster in practice than the current state-of-the-art in trajectory clustering, in some examples by a large margin.
\end{abstract}


\section{Introduction}
For many years, scientists have gathered and studied trajectory data
to understand trends in movement patterns. Ecologists study
animal movements to learn about population growth, social
interactions, as well as feeding and migratory patterns
\cite{brillinger2004,ferreira2011birdvis}. Biologists and computer
scientists study the spread of biological and electronic viruses
\cite{nature2004,nature2007}. Meteorologists use trajectory data to
help predict storm paths \cite{elsner2003tracking,camargoPart1,camargoPart2}, and researchers
from a wide variety of fields study human mobility to perform targeted
advertising, predict traffic and commuting patterns, and data-driven
 urban planning~\cite{becker2011tale,bayir2009discovering}.

The recent ubiquity of GPS and RFID devices has caused a rapid
increase in the amount of available trajectory data. These devices have 
been used to determine locations of animals, shipping containers and different vehicles.
Even in the absence of explicit tracking devices, crowdsourcing can be used 
as an alternative to gather similar data~\cite{ferreira2011birdvis}, 
although with considerable labor requirements.
Another option involves looking at cellular phone handoff patterns:
the traces of calls as they are handed from one cellphone tower to
another~\cite{Becker:2011:RCU:2030112.2030130,pu2011visual,bayir2009discovering}.
This approach can greatly simplify and automate the data acquisition
while still providing complete anonymity for individuals.
In all such cases, due to the vast amount of data
being collected, there is a great need for scalable and efficient
techniques for analyzing this data and discovering the underlying
patterns \cite{gudmundsson2012computational}. 

The analysis of this kind of data is challenging not only because of
its size, but also due to its complexity~\cite{rinzivillo2008visually}.
Trajectories are spatio-temporal
in nature, involving geometric positions, directions, velocities,
durations, life spans, and potentially many other characteristics
specific to the entities being tracked.  
Hurricane tracks may include overall storm strength, wind speeds, or seasonality.  
Animal tracks may be influenced by their size, age, or gender. 
The same attributes can be included for human mobility, as well as their
mode of transportation.
Incorporating these characteristics, when available, can help direct
the trajectory analysis, but also adds complexity.  Gudmundsson et al.
\cite{gudmundsson2012computational} suggest that it may be possible to
infer the characteristics simply by looking at the trajectory data
itself.


In this work, we present a model-based trajectory clustering approach
that uses vector fields as the models for the clustering.  Our method,
called \vfkm, consists of finding vector fields whose integral lines
approximate the given trajectory dataset. The use of vector fields
allows us to naturally encode features of the trajectories such as
direction and speed, which has not been achieved by previous
techniques that used either distance metrics between trajectories or
density-based methods
(\cite{gudmundsson2012computational,lee2007trajectory}).
Our modelling also has the advantage of producing a simple and
physically reasonable modelling.  This is obviously useful when
dealing with datasets representing natural phenomena like storm
tracks (see expermiental restults in
Section~\ref{sec:experiments:hurricane}); however, we also show that
this approach can successfully mine human mobility patterns from GPS
coordinates and even from extremely noisy datasets such as call detail
records (Sections~\ref{sec:experiments:geolife} and
\ref{sec:experiments:cdr}).
Furthermore, vector fields are a good summary for trajectory clusters
and can be easily visualized using any of the numerous techniques
available from the vast literature on vector field visualization.
Previous clustering methods use ``representative'' trajectories as a
way to summarize the result of the clustering process
\cite{lee2007trajectory}, but this is not enough to show all the
possible variability inside a cluster, as demonstrated in
Fig.~\ref{fig:summary_comparison} (see also
Section~\ref{sec:experiments}).

One final argument for clustering a set of trajectories using our
method is its innate capability of handling \emph{partially collected
  or missing data}, a problem that, to the best of our knowledge, has
not been addressed in the literature.  In many situations, it is not
possible to track an entity, thereby generating its trajectory,
throughout its entire lifetime.  For example, in the case of visually
tracking migratory birds, our access might be limited to a very short
time frame. As presented in Section~\ref{sec:experiments:cdr}, some
data, such as call detail records, can only be used in an anonymous
fashion for privacy reasons.  In this case, a trajectory can only be
inferred by the cell tower handoff during an active call and there is
no id that links calls from the same individual. Since users spend
most of their time not making calls, tracks are necessarily partial.
Clustering methods which use representative trajectories to
reconstruct overall patterns will have to resort to
stitching~\cite{lee2007trajectory}. We argue that the technique we
develop here is more natural, and we provide experimental evidence
that it scales favorably. \nivan{could/should we add a section with
  experiments where we throw away some of the segments in the
  hurricane dataset for example just to confirm this?}

For computational efficiency, we use linear approximations of the
trajectories and piecewise linear vector fields as models. This allows
us, as we show in Section~\ref{sec:algorithm}, to define the
trajectory clustering problem as a (constrained) quadratic
minimization problem, that surprisingly connects techniques from
scalar field design on meshes
\cite{sorkine2004least,sorkine2005geometry} and $k$-means clustering.
This results in a very easy to implement and efficient algorithm, as
we demonstrate with several experimental results.  We also compare our
algorithm with state-of-the art trajectory clustering algorithms.
In summary, our contributions are:

\begin{figure}[t]
\centering
\includegraphics[width=\linewidth]{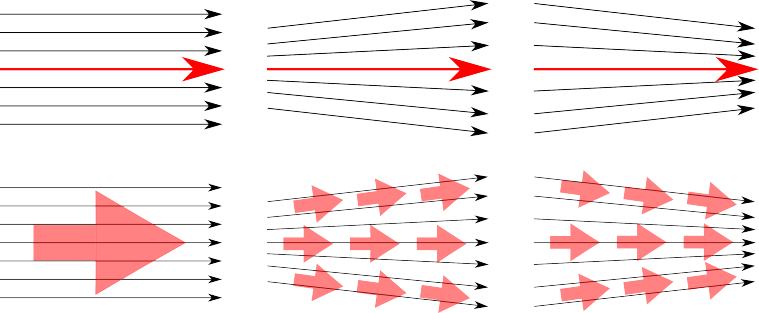}
\caption{\label{fig:summary_comparison} Comparison of summarization of
  a trajectory dataset using representative trajectories and vector
  fields. In the top row, different datasets (black) are summarized
  by the same (red) trajectory, so no variability of the data is
  captured by the summary. In the bottom row, the same datasets are
  summarized using vector fields.  Notice how this method can better
  represent the trajectories in the datasets.}
\end{figure}

\begin{itemize}
\item A novel model-based trajectory clustering method based on vector fields,
called \vfkm, which gives both a partition of trajectories into
meaningful clusters and a best-fitting vector field for each of
them.

\item An experimental analysis of \vfkm through a collection
  of datasets of increasingly large scale, together with a discussion and comparison
  of the results within the context of the current state-of-the-art.
\end{itemize}

In the following sections, we review related work,
introduce the \vfkm technique and show how it can be used in the analysis
of trajectory data.  
We present experimental evidence of its effectiveness using several
data sets, including comparisons to previous methods.
In both synthetic and real
trajectory data, our method can discover significant movement
patterns.

\begin{figure}
\centering
\includegraphics[width=\linewidth]{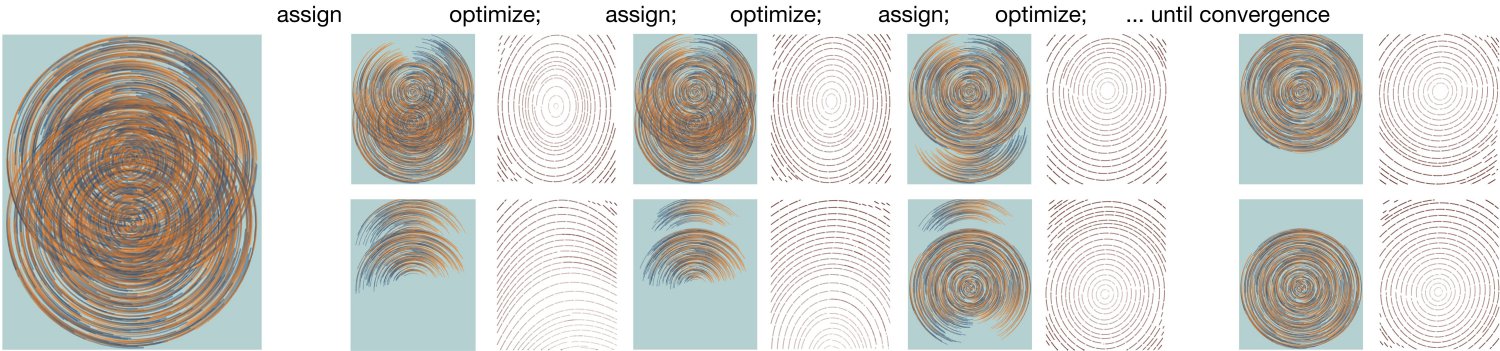}
\caption{\label{fig:overview} An illustration of \vfkm as it partitions 2000 synthetic
  trajectories into two clusters. 
  The algorithm
  alternates between fitting the best possible vector fields from the
  current assignment (``optimize'') and matching trajectories to the
  vector field which fits them best (``assign''). 
  As we show in
  Section~\ref{sec:algorithm}, this procedure always converges. In
  Section~\ref{sec:experiments}, we show that the partitions and the
  vector fields generated by \vfkm encode useful trajectory
  patterns.
  Note that although no individual trajectories
  form a complete circle, \vfkm still recovers the two separate
  circular patterns. 
  The vector fields used in this experiment are linearly
  interpolated over a regular 3x3 grid.
  Our current implementation of the algorithm converged in 30
  iterations on this dataset in about 2 seconds.  }

\end{figure}

\section{Related Work}\label{sec:related_words}


Due to the growing rate at which mobility data is being collected,
\emph{computational movement analysis} is becoming a very active
research field, combining techniques and expertise from many other
fields, including GIS, information visualization, computational
geometry, databases, and data mining
\cite{gudmundsson2012computational}.  In this work, we focus on just
one of the problems in movement analysis, that of extracting arbitrary
movement patterns from trajectory data. In other words, given a large
number of trajectories of moving objects, e.g. animals, people, or
vehicles, we want to quickly identify underlying patterns that exist
and that shed light on the global movement trends of the moving
objects.  As previously described, scientists from many fields can
derive immediate benefits from such trends
\cite{brillinger2004,nature2007,elsner2003tracking}.
Our approach for identifying these patterns
is to perform trajectory clustering.
As Kisilevich et~al.~\cite{kisilevich2010spatio} 
provide a thorough examination of many trajectory clustering techniques, 
we briefly review the most relevant methods here.

Rinzivillo~et~al.~\cite{rinzivillo2008visually} designed a progressive
clustering technique that can utilize different distance functions at
each step of their clustering.  This allows analysis of objects with
heterogeneous properties to be handled differently during the cluster
refinement stages.  Their clustering algorithm is density-based, which
is robust to noise and outliers, a common problem with trajectory
data.  The entire process is visually driven, and interpreting the
(quality of the) clusters involves a human analyst.   
Lee et~al.~\cite{lee2007trajectory} also use density-based clustering, but
believe that clustering whole trajectories may miss common
sub-trajectories. Therefore, they have created the
partition-and-group framework, where they partition the trajectories 
into line segments based on a simplification algorithm and cluster these 
segments using the notions of neighborhood and density. These algorithms
consider trajectories as a set and do not consider the parametrization
of the trajectory and hence they cannot consider information about the
speed in their model. Both methods \cite{rinzivillo2008visually,lee2007trajectory}
rely on the definition of a distance measure between trajectories (or simplified versions
of trajectories such as line segments) which is known to be a difficult problem in the
sense that no proposed distance measure captures well all the attributes of trajectories
\cite{gudmundsson2012computational}. For example, neither of these methods use timing
information for clustering, they only use the geometry of the trajectory and therefore
they cannot enconde speed information what might be relevant in cases like storm track
analysis.
Pelekis et~al.~\cite{pelekis2009clustering} exploit local similarities of
subtrajectories too, but they also study the effect of uncertainty,
e.g. in sampling or in measurement, in the original trajectory data.

Like Rinzivillo~et~al., our overall approach falls within the broader
category of {\it visual and exploratory movement analysis}, which
exploits humans' ability to visually detect patterns, and then steer
the visualization and analysis to those regions of greatest interest.\nivan{is it?}
Andrienko and Andrienko
\cite{andrienko2007,rinzivillo2008visually,andrienko2009} have lead
the field in this area. Their work has focused on human-in-the-loop
analysis systems, but has also included more general aggregation and
visualization of movement data \cite{andrienko2008}, and most recently
the identification of important locations and events by analyzing
movement data \cite{andrienko2011,andrienko2011tvcg}.

Liu et~al.~\cite{liu2011vast} also present a visual analytics system
for exploring route diversity within a city, based on thousands of
taxi trajectories.  Their system offers global views of all
trajectories, but also drills down to routes between
source/destination pairs, and even to specific road segments.  Their
work is more about examining trajectories and less about clustering
them.

An important problem related to trajectory clustering is how to
ultimately visualize trajectory data.  Traditionally flow maps
\cite{phan2005flow,verbeek2011,andrienko2011tvcg} have been used to
convey the amount of people and goods that moved between locations but
without necessarily reporting the exact routes that were taken.  More
recently, there have been several compelling techniques based on density maps
\cite{willems2009,scheepens2011,scheepens2011composite}, and kernel
density estimation \cite{daae2011interactive}.

Vector fields have been widely used in scientific visualization and
even by some researchers doing trajectory clustering analysis to show
speed and direction of animal movements \cite{brillinger2004} and wind
\cite{camargoPart2}.  In these cases, they have only been used to
visualize the results, rather than as an integral part of the
underlying clustering technique.

One important class of clustering frameworks is the model-based clustering
approach in which our method is included.
%
%
The algorithms in this class will typically define a generative
model for the trajectories and then use Maximum Likelihood estimation to fit
the model with the given data.  An important feature of these
algorithms is that it they produce interpretable models for each cluster.
One example of this approach is
the work by Gaffney and Smyth \cite{gaffney1999}, in which they used regression 
mixture models to find their clusters.  A similar idea was used by Wei
et al. \cite{wei2011parallel} in which they used polynomials as models for
the trajecotries.

The idea of using vector fields as tools to analyze trajectory data
was previously used in the image processing community
\cite{nascimento2009,marques2011fast}, but these works used a probabilistic
generative model for trajectory datasets using vector field in which a
trajectory can be generated by different clusters, which can be hard for
an analyst.  In our work, we take a geometric approach to the problem
with no probabilistic assumptions on the data.  As discussed in more
details later, our approach conceptualizes the input data set
as composed of streamlines of a certain number of vector fields.  We
approximate this by considering only piecewise linear vector fields, which
enables us to define the Vector Field \km algorithm on a least
squares optimization that follows the same pattern as the one proposed
in the geometry processing community for triangle mesh
optimization~\cite{sorkine2004least,sorkine2005geometry,nealen2006laplacian}.

\begin{figure}
  \centering
\includegraphics[width=0.9\linewidth]{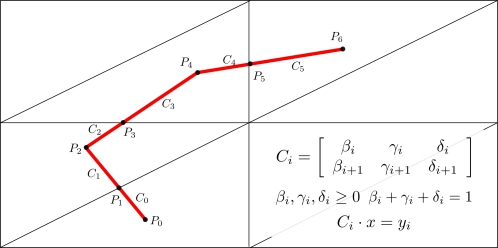}
\caption{This figure illustrates the tessellation of the trajectories,
  so each segment is contained on a face of the grid. The points with
  odd indices, where the original trajectories cross face edges, are
  added in the beginning of the algorithm. In this case each segment
  of the trajectory determines a constraint in the form of a matrix
  $C$.}
\label{figs:constraints}
\end{figure}

\section{Vector Field K-Means}\label{sec:method}

In this section we give a high-level overview of the \vfkm technique.
We start by setting the terminology to be used in the rest of the paper.

\subsection{Terminology}

Trajectories are modeled as paths of the form $\alpha:[t_0,t_1]\rightarrow \mathbb{R}^2$. 
We assume we are given a set of $n$ trajectories $\mathcal{T}=\{\alpha_1,...,\alpha_n\}$. 
Trajectories are given as samples, {\it i.e.}, for each $i=1,...,n$, we are given
a sequence of space-time points \hbox{$\widehat{\alpha_i} = \{(\alpha_i(t^i_1),t^i_1),
(\alpha_i(t^i_2),t^i_2), \hdots, (\alpha_i(t^i_{p_i}),t^i_{p_i})\}$}. 
We approximate each trajectory $\alpha_i$
with piecewise linear curves (constant velocity between two consecutive samples).
This results in a polygonal line representation for each trajectory. 
For each $\alpha_i$ we denote the interval $[t^i_1,t^i_{p_i}]$ by $I_i$ and by $|I_i|$ 
the timespan for $\alpha_i$, {\it i.e.}, $|I_i| = t^i_{p_i} - t^i_1$.
We call each portion of trajectory between two samples as a segment of the trajectory $\alpha_i$. 
For each segment $s_j = [\alpha_i(t_j), \alpha_i(t_{j+1})]$
of $\alpha_i$ we define $\omega_{s_j} = \frac{t_{j+1} - t_{j}}{T}$,
where $T = \sum\limits_{\alpha_i \in \mathcal{T}}^{} |I_i|$ is the total time 
span in the dataset. 

In this paper, our vector fields are always steady and defined
on a domain $\Omega \subset \mathbb{R}^2$, \ie{} they are
functions $X:\Omega\rightarrow \mathbb{R}^2$. We discretize
$\Omega$ as a regular grid $G$ with resolution
$R$ ($R^2$ vertices) and assume linear interpolation
within each face of the grid for the reconstruction of the vector field. We assume that all given trajectories are contained in the domain of 
interest, and also we assume that all the trajectories are tesselated by the grid so that each trajectory
is comprised of segments (portions of the curve between two consecutive vertices) that
do not cross the boundaries of the domain triangles as in Fig.~\ref{figs:constraints}.
Finally, for each segment $s$ of a trajectory $\alpha$ we denote by $C_{s}$ the $2\times R^2$
matrix that contains in the first and second row respectively the baricentric coordinates of the
first and second vertex of the segment  (it has 0 in the entries corresponding to vertices not
contained in the face where the segment $s$ lies), Fig.~\ref{figs:constraints} illustrates the setting.
\nivan{I moved the comments on the trajectories tesselation and C matrix to this section, still need to 
remove them from the rest of the text. Still need to verify that this is enough.}

\subsection{Method}
Our ultimate goal is to mine movement patterns within large trajectory
datasets. Our approach is to capture those patterns by defining a vector field for which the
trajectories are approximately integral lines, according to a reasonable error measure.
However, any non-trivial dataset is likely to contain
trajectories that cannot be modeled as streamlines of a \emph{single} vector
field. We use this fact to define a similarity notion between
trajectories, namely how well these trajectories can approximate streamlines
of a single vector field. We then propose using this similarity to find multiple 
vector fields that capture the movement features of the dataset under consideration.
Vector field \km attempts to separate the trajectories into a small
number of clusters according to the best vector field that approximates them.

More formally, the basic assumption of our approach is that for every
set of trajectories $\mathcal{T}$, there exists a set of reasonably
smooth vector fields $X_i \in F, |F| = k$ that explains most of the
mobility in the data, in the sense that each trajectory would be
approximately tangent to one of the $X_i$. As discussed in the
previous paragraph, in general we must have $k > 1$. Therefore we see
each vector field not only as a summary of the trajectory cluster, but
also as the center of the cluster, and thus analogous to the original
\km algorithm \cite{kmeans}.

The problem then is to (1) define these vector fields and (2) assign each
trajectory to the vector field that fits it best. In other
words, we need to compute both the best-fitting vector fields and a
function $\Phi: \mathcal{T} \to \{1, 2, \cdots, k\}$ which assigns the
trajectories to the best vector field. We propose to look for
reasonably smooth vector fields that are approximately tangent to
the trajectories. We propose an iterative process that uses the results of step
(2) in order to perform step (1) and conversely uses results from step (2) 
to perform step (1). More clearly, the Vector field \km algorithm 
consists of the following two steps
\begin{itemize}
\item[(*)] Given any candidate assignment of trajectories $\Phi$, for 
each set $\Phi^{-1}(i),$ $i=1,\hdots,k$, we find the best-fitting
vector fields \emph{over those particular trajectories}, and
\item[(**)] Given a set of vector fields $\tilde{V} = \{X_1,\hdots,X_k\}$, in order to
compute the best assignment function \emph{for those particular vector
  fields} we simply evaluate the error (defined later) over each vector
field, and pick the best field for each trajectory.
\end{itemize}

Algorithm~\ref{alg:outlineCode} contains the outline of the Vector field \km algorithm.
In this pseudo-code, the step $(*)$ corresponds to the $fitVectorField$ routine and
as we will show below, we can formulate this step as a linear system whose solution can be
computed essentially in linear time, and which gives us the smoothest,
best-fitting vector field for a set of trajectories. As mentioned earlier, this system
has the same general form as the ones proposed by \cite{sorkine2004least,sorkine2005geometry,nealen2006laplacian} for polygonal mesh processing. The step $(**)$ corresponds to 
finding the vector field with smallest error with respect to a given trajectory. 
The error measure $E$ is going to be defined
later. We highlight that although vector field \km follows the same lines of the
ordinary \km \cite{kmeans} algorithm, they are fundamentally different
in the sense that the cluster ``centers'' are objects of a
\emph{different nature} than the objects being clustered.

\begin{figure}
  \centering
\includegraphics[width=0.9\linewidth]{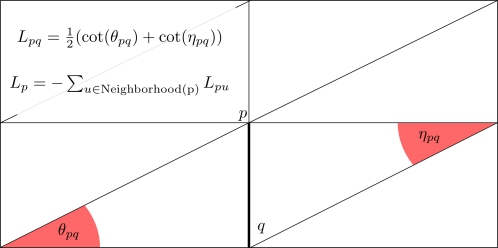}
\caption{This figure illustrates the computation of the Laplacian matrix
  that we use as a smoothness penalty, to favor simpler vector fields
  over more complicated ones.}
\label{figs:laplacian}
\end{figure}


The execution of \vfkm on a synthetic dataset with 2000 overlapping
trajectories is illustrated in Fig.~\ref{fig:overview}.  In each
iteration of the algorithm, the two clusters of trajectories and their
corresponding vector fields are improved until a perfect separation
occurs at iteration 30. Section \ref{sec:algorithm} explains the
algorithm in detail. 

\begin{algorithm}
\caption{Vector Field K-Means Outline}
\label{alg:outlineCode}
\begin{algorithmic}
\REQUIRE $k$: Number of clusters,  $\mathcal{T} = \{\alpha_1,\hdots,\alpha_n\}$: Array of curves 
\ENSURE $V=\{X_1,\hdots,X_k\}$, $\Phi:\mathcal{T}\rightarrow\{1,\hdots,k\}$
\STATE $\Phi \leftarrow $ Initialize($\mathcal{T}$,$k$)
\REPEAT
     \FOR{$i = 1$ to $k$}         
         \STATE $f_i \leftarrow$ $\fitVectorField(\Phi^{-1}(i))$
     \ENDFOR\\
     \FOR{$i = 1$ to $n$}
         \STATE $j_0 \leftarrow \underset{j \in \{1,2,...,k\}}{\operatorname{argmin }} E(X_j, Curves(i))$
         \STATE $\Phi(\alpha_i) \leftarrow j_0$
     \ENDFOR
\UNTIL{not converge}
\end{algorithmic}
\end{algorithm} 



\section{Algorithm}\label{sec:algorithm}

In this section we describe the algorithm in
detail. \VFKM finds a minimum of the following energy:
\begin{equation}E = \underset{X_j, \Phi}{\min} \, \lambda_L \sum_{j=1}^k
  \left|\left|\Delta X_j\right|\right|^2 + \sum_{\alpha_i \in \Phi^{-1}(j)} \int_{t^i_1}^{t^i_{p_i}}
  \left|\left|f(\alpha_i(t)) - \alpha_i'(t)\right|\right|^2 \, dt
\label{eq:formulation}\end{equation}

This energy is always non-negative. As we will show, each
iteration of \vfkm reduces E, and since there's only a finite number
of assignments, \vfkm is guaranteed to terminate.
In equation~\ref{eq:formulation}, $\lambda_L$ plays the role of a
weighting factor: if close to
$0$, we are giving relatively high priority to 
the trajectory constraints through the matrices $C_s$. When it is
close to $1$, solutions will tend to be smoother, since most of the
energy is spent minimizing $\sum_i||\Delta X_i||^2$. 


\subsection{Fitting Vector Fields}

The $fitVectorField$ routine is the central step of \vfkm.  It
consists of an optimization problem with two types of constraints:
value and smoothness, defined below.  As depicted in
Algorithm~\ref{alg:outlineCode}, in this step we are given a subset
$\mathcal{T'}$ of $\mathcal{T}$.  We formulate the vector field
fitting problem as a lest-squares minimization problem.  To simplify
our discussion, we consider only a single trajectory $\alpha_i\in
\mathcal{T'}$ and show how to build the optimization problem for it.
We then formulate the problem for the entire set $\mathcal{T'}$ by
simply putting together the constraints for the different trajectories
in $\mathcal{T'}$.
We want to get a vector field for which the trajectory $\alpha_i$ is (approximately) 
a integral line, {\it i.e.}, we must have $X(\alpha_i(t)) = \alpha_i'(t)$. We see each
$t \in I_i$ as a contraint to the value of $X$. In our discrete setting, 
given $\alpha_i(t)$ we denote by $C_{\alpha_i(t)} X$ the line vector 
indexed by the vertices of the grid containing
the baricentric coordinates of $\alpha_i(t)$ (it has 0 in the entries not
contained in the face where $\alpha_i(t)$ lies). The constraint equation just mentioned
becomes $C_{\alpha_i(t)} X = \alpha_i'(t)$. 


\subsubsection{Value Constraints: from discrete to continuous constraints}\label{sec:lambda}

We have infinitely many such constraints, so we cannot directly write
the constraints for all the points on $\alpha_i$. In this section we
show that we can write all this contraints in a matricial form on the
segments of $\alpha_i$. Consider each segment $s$ of $\alpha$
and denote by $C_s$ the ($2\times R^2$) matrix containing the baricentric coordinates of the
endpoints of $s$. Furthermore consider the ($m+1\times 2$) matrix $\Lambda_m$:
 
\[\Lambda_m = \frac{1}{\sqrt{m}} \left( \begin{array}{cc}
1 & 0 \\
1-1/m & 1/m \\
\vdots \\
1/m & 1-1/m \\
0 & 1 \end{array}\right).\]

Consider the equation $\Lambda_m C_s x = \Lambda_m \alpha'_s$, where $\alpha'_s$
denotes the vector containing the velocity vectors of $\alpha$ at the endpoints
of $s$. This equation simply represents the process
of sampling $m+1$ equally spaced points on the segment $s$ and writing down
the constraints as discussed in the previous section. More clearly, if
$m = 1$, we simply have that $\Lambda_1$ is the identity matrix and the
matrices $\Lambda_m C_s$ introduce only endpoint constraints. If $n=2$, on the
other hand, we would have constraints not only on the endpoints but on
the midpoint of the segment as well. The normalization factor $\frac{1}{\sqrt{m}}$
keeps the total weight of this constraint the same as $m$ grows. By increasing $m$, we set
constraints over progressively larger sets of points on each
segment. We are interested in the case where $m$ grows without
bounds. Although we appear to have an arbitrarily large $\Lambda_m$
matrix and hence an arbitrarily large number of constraints, the
matrix $\Lambda_m$ only appears in the normal equations of the least squares
problem, and so is always multiplied by its own transpose. In that case, $\Lambda_m^T \Lambda_m$ is
always a $2\times 2$ matrix, and so we can define the matrix $\Lambda$ to
be the positive-definite square root of the following limit:
\begin{eqnarray*}
\Lambda^T \Lambda &=& \lim_{n\to\infty}\Lambda_n^T \Lambda^n\\
\Lambda &=& \left ( \begin{array}{cc}
  \frac{1}{2}\left(\frac{1}{\sqrt{2}} + \frac{1}{\sqrt{6}}\right) & \frac{1}{2}\left(\frac{1}{\sqrt{2}} - \frac{1}{\sqrt{6}}\right) \\
  \frac{1}{2}\left(\frac{1}{\sqrt{2}} - \frac{1}{\sqrt{6}}\right) &  \frac{1}{2}\left(\frac{1}{\sqrt{2}} + \frac{1}{\sqrt{6}}\right) \end{array} \right ) 
\end{eqnarray*}
We then (re)define the value constraints for $\alpha_i$ as
$\varepsilon(X,\alpha_i):=\sum_{s \in \alpha_i} \omega_s||\Lambda(C_s X -
\alpha_i^s)||_2^2$, where $\alpha_i^{s'}$ denotes the vector containing the
velocity vector of $\alpha_i$ at the endpoints of $s$. Thus, we get the
interesting property that the linear system we solve simultaneously
tries to satisfy an infinite number of value constraints, while still
working in a finite dimensional setting. Using $\Lambda$, $\varepsilon(X,\alpha_i)$ 
is essentially square of the $L^2$ norm of the difference between the vector field and the
trajectory velocity vector on the trajectory points. In order to see this, we notice that

\begin{eqnarray*}
||X\circ \alpha_i - \alpha_i'||_{L^2}^2 &=& \int_a^b ||X(\alpha_i(t)) - \alpha_i'(t)||dt = \\
\sum\limits_{j=1}^{n-1}\int_{t_j}^{t_{j+1}}||X(\alpha_i(t)) - \alpha_i'(t)||_{L^2}^2 &=& 
\sum\limits_{j=1}^{n-1}\int_{t_j}^{t_{j+1}}||X(\alpha_i(t)) - \alpha_i^{j'}||_{L^2}^2,
\end{eqnarray*}
where $\alpha_i^{j'}$ denotes the velocity vector of the $i^{th}$ segment of $\alpha_i$.
Therefore,
\begin{eqnarray*}
&||X\circ \alpha_i - \alpha_i'||_{L^2}^2 = \sum\limits_{j=1}^{n-1}\int_{t_j}^{t_{j+1}}||X(\alpha_i(t)) - \alpha_i^{j'}||_{L^2}^2 = \\
&\sum\limits_{j=1}^{n-1} (t_{j+1} - t_j) \int_{0}^{1}||(1 - \sigma)X(\alpha_i(t_j)) + \sigma X(\alpha_i(t_j)) - \alpha_i^{j'}||^2 d\sigma = \\
&\sum\limits_{j=1}^{n-1} (t_{j+1} - t_j) \int_{0}^{1}||(1 - \sigma)(X(\alpha_i(t_j)) - \alpha_i^{j'}) + \sigma (X(\alpha_i(t_j))- \alpha_i^{j'})||^2 d\sigma = \\
&\sum\limits_{j=1}^{n-1} (t_{j+1} - t_j)(
\frac{1}{3}\left|\left|X(\alpha_i(t_j))    - \alpha_i^{j'}\right|\right|^2 + 
\frac{1}{3}\left|\left|X(\alpha_i(t_{j+1})) - \alpha_i^{j'}\right|\right|^2 +\\
&\frac{1}{6}( ( X(\alpha_i(t_j)) - \alpha_i^{j'}) \cdot (X(\alpha_i(t_{j+1})) - \alpha_i^{j'}))) = 
T\varepsilon(X, \alpha_i)
\end{eqnarray*}.

\subsubsection{Smoothness Constraint}\label{sec:smoothness_constraint}

As stated before we also include a smoothness regularity constraint on the vector field.  
In order to do so we use the the Laplace-Beltrami operator on a vector field. In order to model it,
we use the well-known matrix representation of the
Laplace-Beltrami operator over polyhedral surfaces and the cotangent
formula \cite{wardetzky2008convergence}.  In general, if we let $p,q$
be a pair of adjacent vertices of the grid, then the entries of the cotangent
Laplacian matrix L are given by
\begin{eqnarray*}
\bar{L}_{pq} = \frac{1}{2}(\cot\theta_{pq} + \cot\eta_{pq}),\\
\bar{L}_p = -\sum\limits_{t \in Neighborhood(p)}\Delta_{pt}.
\end{eqnarray*}
This formula allows us to compute the Laplacian of scalar functions
defined over the grid by simply representing the values of the
function on the vertices of the grid as a vector and multiplying the
vector field by the Laplacian matrix. We illustrate the construction
of the Laplacian matrix in Fig.~\ref{figs:laplacian}.

\subsubsection{Fitting Process}\label{sec:fitting_process}

We want the vector field $X$ to be defined as the solution
of the Laplace equation $L X = 0$ (vector Laplacian) with the
value constraints defined in the previous section. Therefore, we define the best
fitting vector field $X$ for a set of trajectories as the solution (in the least squares sense) 
of the following system:
\begin{eqnarray*}
Lx &=& 0 \\
\Lambda C_s x &=& \Lambda y_s
\end{eqnarray*}
In the above, $s$ ranges over all segments of the trajectories
assigned to the particular vector field. We incorporate the weights on
the associated linear system for the normal equations, and solve the
following linear system with a unique solution:
\begin{equation*}
\left(L^TL(\lambda_L \sum_s \omega_s) + (1-\lambda_L)\sum_s \omega_s (C_s^T \Lambda^T
\Lambda C_s)\right) X = \sum_s \omega_s (C_s^T \Lambda^T \Lambda y_s),
\end{equation*}
where $\lambda_L$ is a parameter that idicates the weight
of the smoothness penalty in the optimization problem. 

Finally we notice that this optimization problem has the same form as the one presented in
\cite{sorkine2004least,sorkine2005geometry} used in that context for scalar field design
on meshes.

\subsection{Assigning trajectories to vector fields}\label{sec:assign_step}
In the second phase of the Algorithm~\ref{alg:outlineCode}, we assume we have a fixed set of vector fields
$\mathcal{V} = \{X_1,\ldots,X_k\}$ and a set $\mathcal{T} =
\{\alpha_1,...,\alpha_n\}$ of trajectories. 
The goal is to build the next function $\Phi$ that assigns each trajectory to one of the $k$ cluster
centers, \ie the vector fields $X_1,\hdots,X_k$. 

The assignment algorithm is trivial: for each
vector field $X_i$ and trajectory $\alpha$, we simply evaluate
\[\varepsilon(X_i, \alpha)(1 - \lambda_L) + ||LX_i||^2\lambda_L.\]
The new assignment is then the global minimizer of the
$k$ possible choices.

\begin{figure}
\centering
{\scriptsize
\begin{tabular}{|c|c|c|c|c|c|c|}
\hline
dataset & \#traj. & res. & $k$ & fit & eval & assign \\
\hline
Synthetic & 2000 & 3 & 2 & 1.730s & 0.076s & 0.074s \\
Atlantic & 1415 & 5 & 7 & 8.076s & 0.176s & 0.335s \\
Beijing Wide & 45563 & 5 & 4 & 110.99s & 2.377s & 4.265s \\
Beijing Campus & 12883 & 10 & 16 & 124.35s & 1.229s & 2.195s \\
CDR & 37435 & 4 & 4 & 201.24s & 4.227s & 7.597s \\
CDR Large & 372601 & 4 & 4 & 2497s & 43.127s & 75.24s  \\
\hline
\end{tabular}}
\caption{Experimental results of \vfkm.  For each dataset, we report
  the number of trajectories, the grid resolution (res), the number of
  clusters (k), and the total running times (in seconds) for the
  vector field fitting routine, the error evaluations, and the
  trajectory assignments.}
\label{tab:timing}
\end{figure}

\begin{figure}
  \centering
\includegraphics[width=1\linewidth]{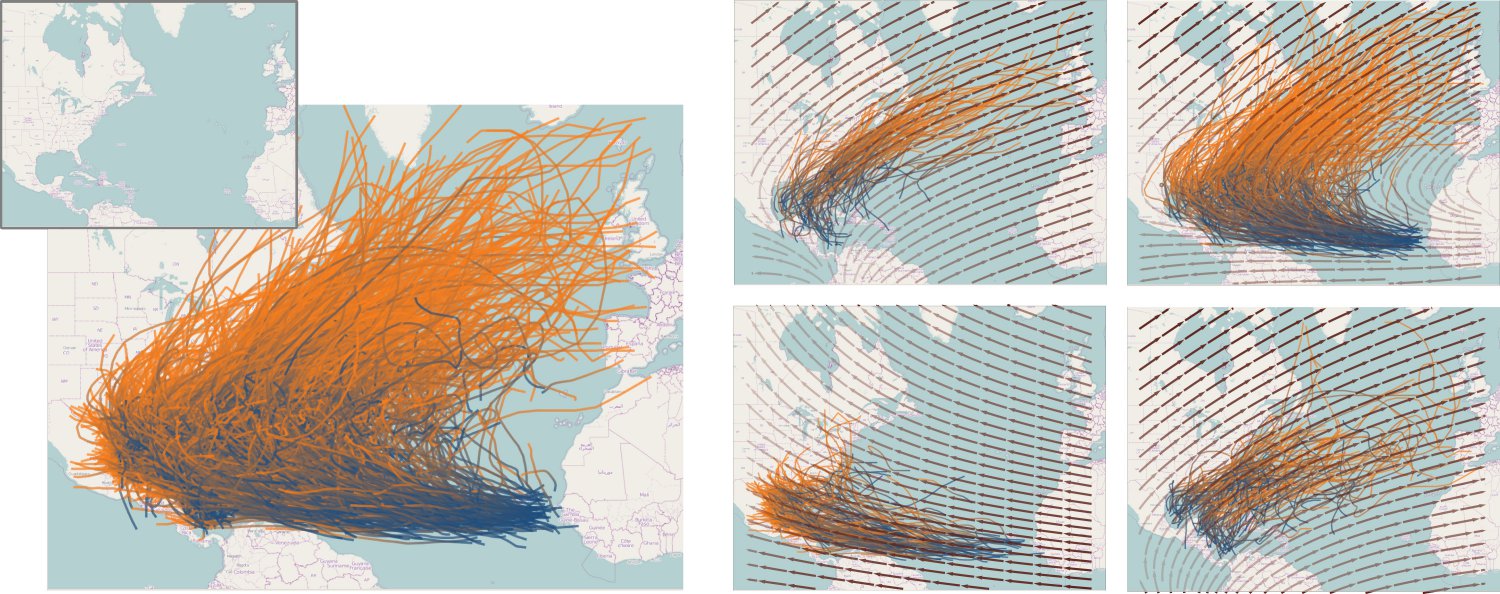}
\caption{Results of clustering trajectories from the HURDAT
  dataset~\cite{hurdat}. On the left, the original input trajectories
  are shown.  On the right, four trajectory clusters and their
  corresponding vector fields showing relative speed: the top-left vector
  field heading northeast is generally faster than the one on the
  bottom-right \jim{Can we see that in the vector fields?  The arrows in our tool would show this, but the overlays do not IMHO}. The direction of each trajectory is shown colormapped
  from blue to orange. }
\label{figs:atlantictrajectories}
\end{figure}

\begin{algorithm}
\caption{Initialization Algorithm}
\label{alg:initializationCode}
\begin{algorithmic}
\REQUIRE $k$: Number of clusters,  $Curves$: Array of curves 
\ENSURE $\Phi$: Initial clusters
\STATE $c \leftarrow $ random element from Curves
\STATE $f_1 \leftarrow$ $\fitVectorField(\{c\})$
\FOR{i = 2 to k}
\STATE $c \leftarrow \underset{c' \in Curves}{\operatorname{argmax }} \{\underset{\{j|1\leq j < i\}}{\operatorname{min }} E(f_{j}, c')\}$
\STATE $f_i \leftarrow$ $\fitVectorField(\{c\})$
\ENDFOR
\FOR{i = 1 to Curves.size()}
\STATE $j_0 \leftarrow \underset{j \in \{1,2,...,k\}}{\operatorname{argmin }} E(f_j, Curves(i))$
\STATE $\Phi(i) \leftarrow j_0$
\ENDFOR
\end{algorithmic}
\end{algorithm} 

\subsection{Algorithm Initialization}\label{sec:initialization}
As with traditional \km, our initialization step has a clear influence
on the final results.  We implemented a simple method to choose the
initial vector fields and trajectory partitions that was effective in
our experiments.  The main idea is to try to have as diverse initial
clusters as possible (see pseudo-code in
Algorithm~\ref{alg:initializationCode}).  The algorithm takes as
inputs an array of curves and a number $k$ of clusters to be created,
and starts by choosing a trajectory $\alpha: [t_0, t_1] \to R^2$ at
random to be part of the first cluster.  It uses the $fitVectorField$
routine previously described to fit a single trajectory to the first
vector field. The algorithm proceeds by fitting to the $i$-th vector
field the trajectory that has the worst error among all previously fit
vector fields. After computing $k$ vector fields, we compute the
assignment $\Phi$ by picking the best vector fields for each
trajectory.

\subsection{Computational Complexity and Implementation}

As discussed in Section~\ref{sec:assign_step}, the assignment step
consists of a linear pass over the trajectory data, and for each of
these, we need to find the vector field that minimizes the error
(defined in Section~\ref{sec:assign_step}). This can be implemented in $O(k
|S(\mathcal{T})|)$, where $S(\mathcal{T})$ denotes the set of line
segments that compose the trajectories in $\mathcal{T}$.

For the first step, we have used a simple Unconstrained Conjugate
Gradient \cite{shewchuk1994introduction} algorithm as a linear system
solver. Therefore, the complexity of this step is
given by $O(kN(R^2 + |S(\mathcal{T})|))$, where $N$ denotes the
maximum number of iterations of the Conjugate Gradient Method, $R^2$
denotes the grid resolution corresponding to the multiplication by the
Laplacian matrix, and $|S(\mathcal{T})|$ corresponds to the
multiplication by the constriant matrix $C$. As we see in the experiments
good results can be obtained with relative low values of $R$ and hence
the complexity is dominated by $kN|S(\mathcal{T})|$. Notice that the
algorithm is highly parallelizable (although
we did not take advantage of that in our current implementation), 
the {\it fitVectorField} procedure can be executed independently for each cluster.

The choice of the Conjugate Gradients solver was made simply out of
convenience; however, we can further optimize our current
implementation by using more sophisticated methods to solve systems of
linear equations, e.g. Constrained Conjugate Gradients
\cite{shewchuk1994introduction} and Cholesky Decomposition
\cite{trefethen1997numerical}. 

\begin{figure}[t]
\centerline{\includegraphics[width=11cm]{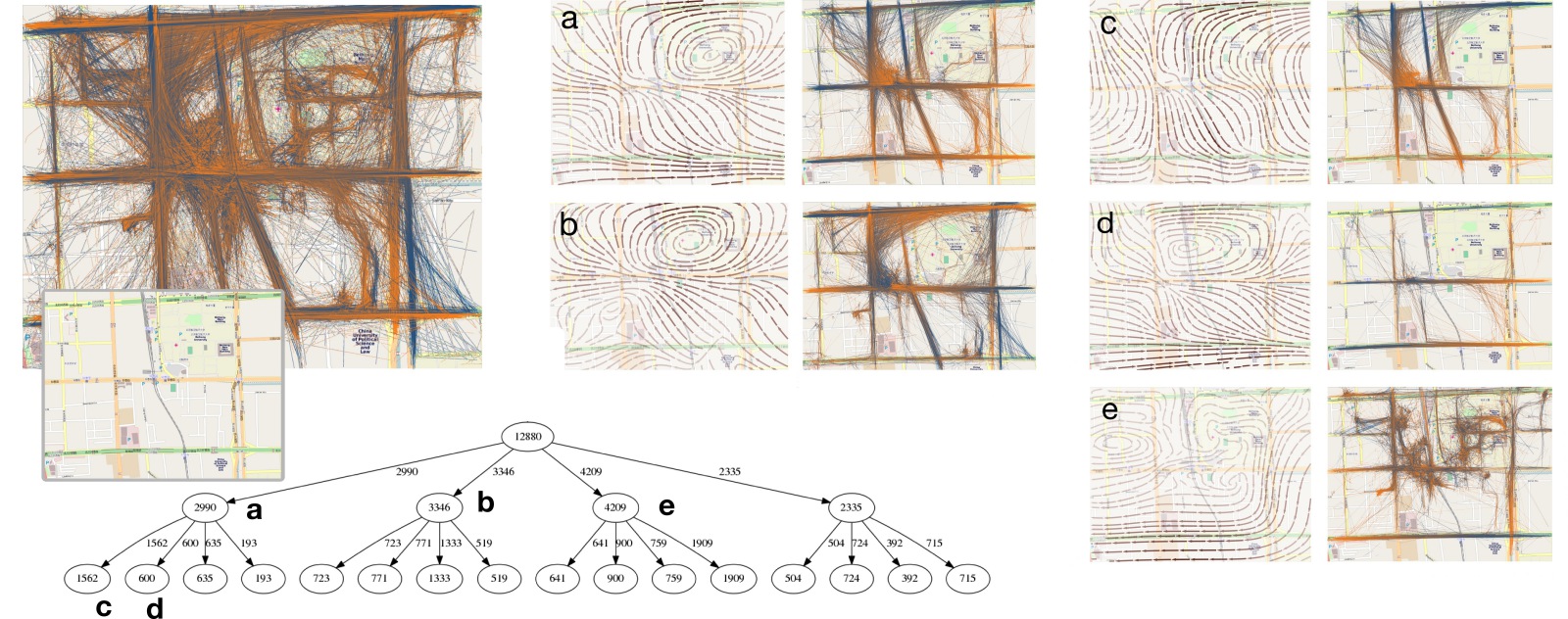}}
\caption{This figure shows about 13000 trajectories from the GeoLife
  Trajectories dataset clustered using \vfkm.
  The original trajectories were cropped outside of a small area, and
  sampled to a 2-minute per-sample resolution.
  The direction of each individual trajectory is shown colormapped from blue to orange.
  In the inset, we show the corresponding OpenStreetMap tile for
  this area of Beijing together with a dendrogram of the
  clustering. 
  We partition the original data
  into four clusters, and then partition each cluster a second time,
  resulting in 16 subclusters.  
  Images (a), (b), and (e) illustrate three of the
  original clusters together with their corresponding vector fields,
  while images (c) and (d) show two of the resulting subclusters.
  Using \vfkm, we can clearly identify movement patterns,
  including separating faster vehicular (a, b) from
  slower pedestrian traffic (e).}    
  \label{figs:teaser}
\end{figure}

\section{Experiments and Results}\label{sec:experiments}

We now report the results of running \vfkm.  In our experiments, the
algorithm was able to efficiently extract significant movement
patterns across diverse datasets.  We start with a synthetic dataset,
and progressively move up to larger examples.  Our final result
involves clustering over 370,000 very noisy trajectories.  All
reported running times (see Fig.~\ref{tab:timing}) are from our
prototype implementation: a single-threaded, single-process C++
application running on an Intel Core i7-960 desktop with 6GB of
RAM. The total amount of memory required by our application remained
under 1GB for all reported experiments.

\begin{figure}
  \centering
\includegraphics[width=\linewidth]{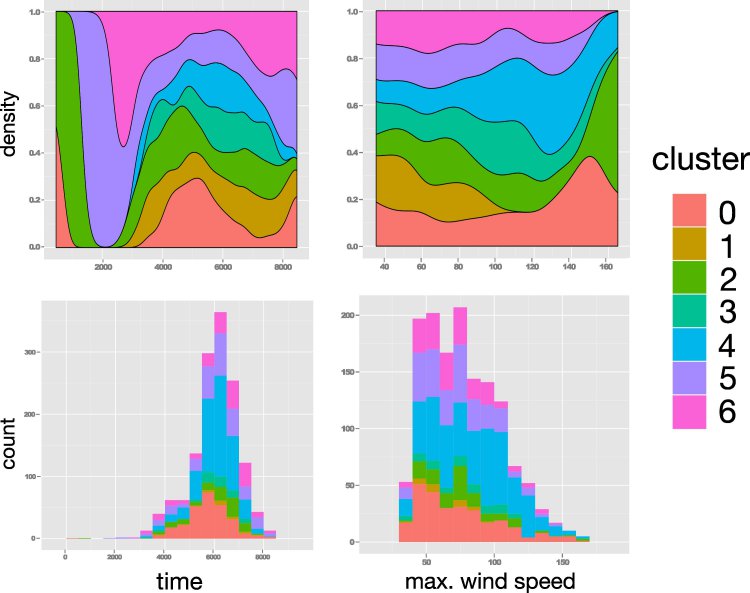}
\caption{Time (measured in hours since the beginning of the year) and
  maximum wind speed (miles per hour) distributions for Atlantic
  hurricanes. Compared to cluster 4, cluster 0 storms tend to happen
  earlier in the year.  Stronger storms appear also more likely to be
  in cluster 0, although there are few samples in that region. The
  wide density bands in clusters 2 and 5 are due to outlier storms in
  strength and maximum wind speed. These attributes are not taken into
  account for clustering; the features appear simply from the tracks
  in clusters having related attributes.  }
\label{figs:atlanticcharts}
\end{figure}

\subsection{Synthetic Data}

In this synthetic dataset, we assume the existence of two overlapping
circulatory movement patterns.  Each trajectory covers a partial,
randomly selected section of the circle at a random distance from the
center. We sample 1,000 trajectories from each overlapping pattern.
As we show in Fig.~\ref{fig:overview}, \vfkm recovers the two
overlapping patterns perfectly. This shows, very clearly, that \vfkm
does not create clusters by selecting representative trajectories at
all: its vector fields fit \emph{all} circular trajectories equally
well. \jim{Perhaps a little more exposition to hammer home our points
that reviewers missed???}

\subsection{Atlantic Hurricanes}\label{sec:experiments:hurricane}

\begin{figure}
\centering
\includegraphics[width=\linewidth]{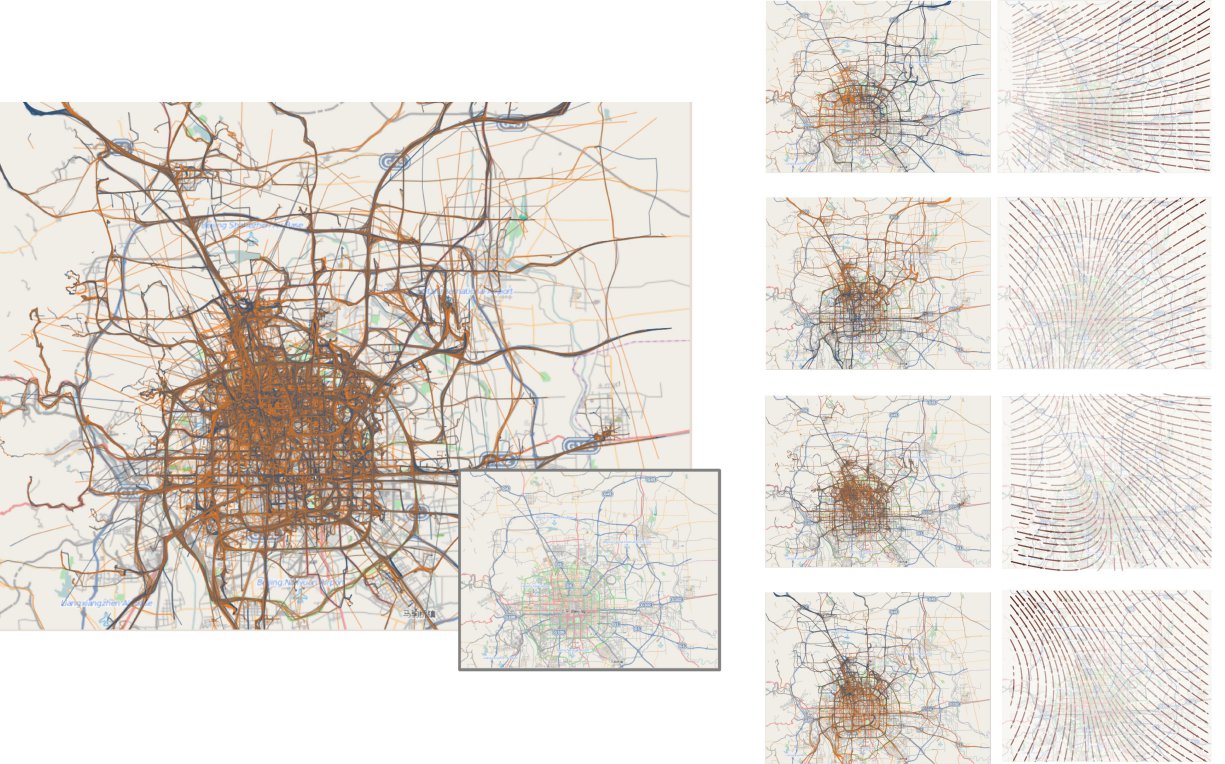}
\caption{Large-scale movement patterns around Beijing, GeoLife Trajectories
  dataset. All but one cluster appear to depict travel in and out of
  the city from the surrounding highways. The remaining
  cluster (third from top) has much slower speeds than the other
  clusters. Its trajectories are also more tightly packed around a
  small region, and this led us to the experiment shown in Figs.~\ref{figs:teaser} and
 \ref{figs:lunch}.
}
\label{figs:beijing_coarse}
\vspace{-1.5em}
\end{figure}

HURDAT is a hurricane tracking dataset maintained by the National
Hurricane Center (NHC)~\cite{hurdat}. The dataset contains 1415
trajectories of different Atlantic storms between the years 1861 and
2011. It contains not only position and time information, but also
sustained surface wind speeds, and sea-level pressure information. The
data are recorded for every active tropical storm, with a resolution
of 6 hours. For the purpose of \vfkm we only need the latitude,
longitude, and time of each track.

We show four of the seven clusters of our analysis in
Fig.~\ref{figs:atlantictrajectories}. Note that \vfkm separates what
originally looks like a fairly uniform set of trajectories. One of the
clusters neatly capture Cape Verde hurricanes which tend to make landfall in
North America, while two separate clusters show hurricanes which
originate in the Caribbean and Gulf of Mexico. Upon closer inspection,
it appears that this separation of two similar looking trajectory
clusters is due to the more chaotic trajectories of one of the
clusters, which result in a generally lower-velocity vector field. The
remaining cluster has hurricanes with the common trajectory of
crossing the Atlantic and moving north-east along the east coast of
the United States. The clusters we do not show appear to contain
mostly outlier storms, one of which contains 21 hurricanes which move
in a northerly fashion.

Vector field \km clearly captures the movement patterns of these
hurricanes. To assess whether the clusters contain significant other
information, we investigated histograms and conditional probability
distributions of wind speeds and time of hurricane occurrence, shown
in Fig.~\ref{figs:atlanticcharts}.  Attributes that were not taken
into account during the clustering are evident when examining the
trajectories assigned to each cluster. \jim{I have trouble seeing our
claims in the plots to be honest.  We may need to rework them some}

\subsection{GeoLife GPS Trajectory Dataset}\label{sec:experiments:geolife}
\begin{figure}
\centering
\includegraphics[width=\linewidth]{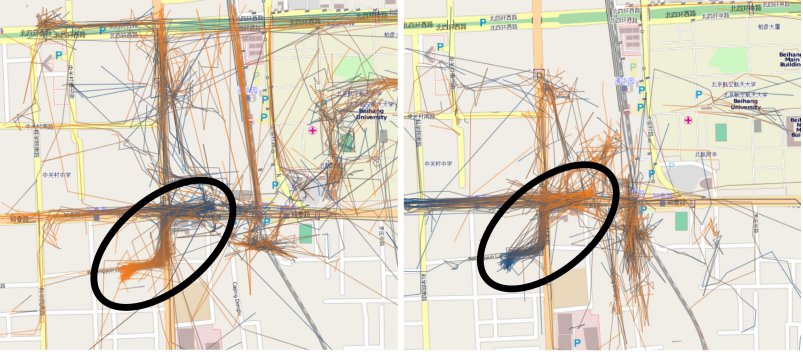}
\caption{Two of the clusters identified by \vfkm seem to include a
  particularly repetitive slow trajectory path, which by map
  inspection we speculate to be a nearby lunch spot for area workers.}
\label{figs:lunch}
\end{figure}
The GeoLife GPS dataset consists of a collection of 17,621
trajectories recorded by Microsoft Research at Beijing. The
trajectories are GPS tracks of 178 users over a three year period from
April 2007 to October 2011
\cite{zheng2008understanding,zheng2009mining,zheng2010geolife}. Although
the entire dataset encompasses trajectories throughout the entire
planet, in this study we focus on two different regions around
Beijing. The raw trajectories from the dataset are unsegmented: some
GPS tracks run for days. We split trajectories with the following very
simple rule: whenever the time between two samples is larger than 2.5
times the median time between samples, we break off the
trajectory. Finally, the dataset is quite densely sampled. We reduce
the sampling rate by only keeping measurements at least 2 minutes
apart from each other.

\begin{figure}[t]
\centering
\includegraphics[width=.9\linewidth]{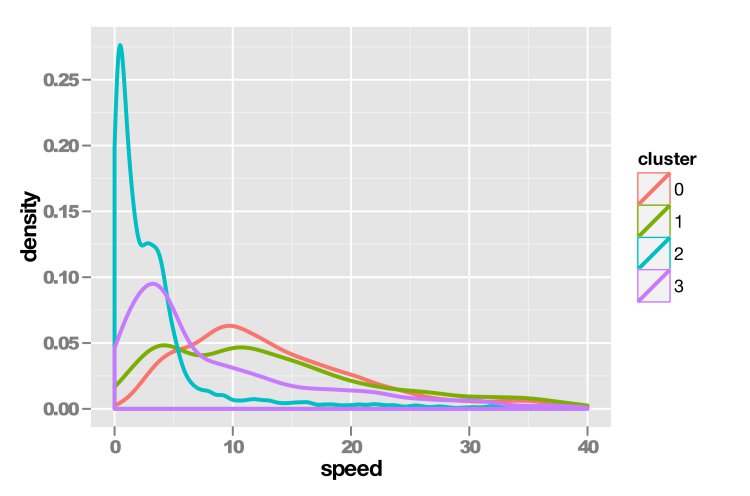}
\caption{In the GeoLife dataset cropped as shown in
  Fig.~\ref{figs:teaser}, different clusters capture
  different speed features of the trajectory data. The clusters shown
  here are the first level of the tree, with respectively 2990, 3346,
  4209 and 2335 trajectories each. Cluster 2 (named ``e'' in
  Fig.~\ref{figs:teaser}) has significantly slower trajectories.}
\label{figs:beijing_speeds}
\end{figure}

Fig.~\ref{figs:beijing_coarse} shows a first run of \vfkm on the
GeoLife dataset in which the algorithm was able to find general
movement trends within the trajectories.  Three of these are clear
directional patterns of trajectories heading west, north, and south.
We speculate these to be mainly commuting patterns, since the fourth
remaining cluster consists essentially of trajectories inside the
city's road network. Although we believe that with the chosen
resolution (5x5) \vfkm cannot reliably resolve the patterns in that
cluster (and hence the vector field is not very informative), the
large density of trajectories around a relatively small area in the
cluster suggested to us further analysis centered in that region.

Fig.~\ref{figs:teaser} shows an exploration we performed on that
narrower region of Beijing. For that same dataset,
Fig.~\ref{figs:beijing_speeds} shows the distribution of average
speed for the first level of the tree in Fig.~\ref{figs:teaser}.
Notice that \vfkm is able to separate trajectories not only by their
overall movement trend (direction), but also by their speed: cluster (e)
clearly contains slower trajectories, which, by examination, seem to be
pedestrian traffic. We also notice that
cluster (b) contains the fastest trajectories in the database.  We found
two intriguing patterns in cluster (e): people apparently going to a
train station and heading to what seems to be a lunch spot. We
highlight this in Fig.~\ref{figs:lunch}.

\subsection{Call Detail Record Dataset}\label{sec:experiments:cdr}

We collected anonymized Call Detail Records (CDR) from the cellular network
of a large US communications service provider. We captured the handoff
patterns carried out by approximately 300 cell towers located in the
vicinity of Anytown, a suburban city with approximately 20,000
residents.  Our goal was to capture handoffs related to
vehicular traffic in and around the town.

\begin{figure}
\centering
\includegraphics[width=\linewidth]{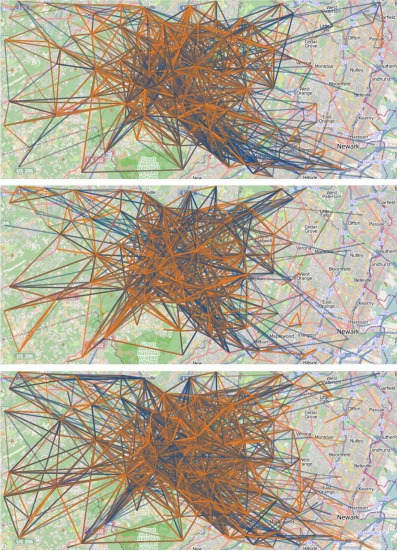}
\caption{\label{fig:handoffs} The anonymized call detail records for
  over 370,000 cell phone calls produced noisy trajectories
  (i.e. hand-offs between cell phone towers) around a suburban city.
  Three of the four clusters computed by \vfkm are shown.}
\end{figure}

\begin{figure}
\centering
\includegraphics[width=\linewidth]{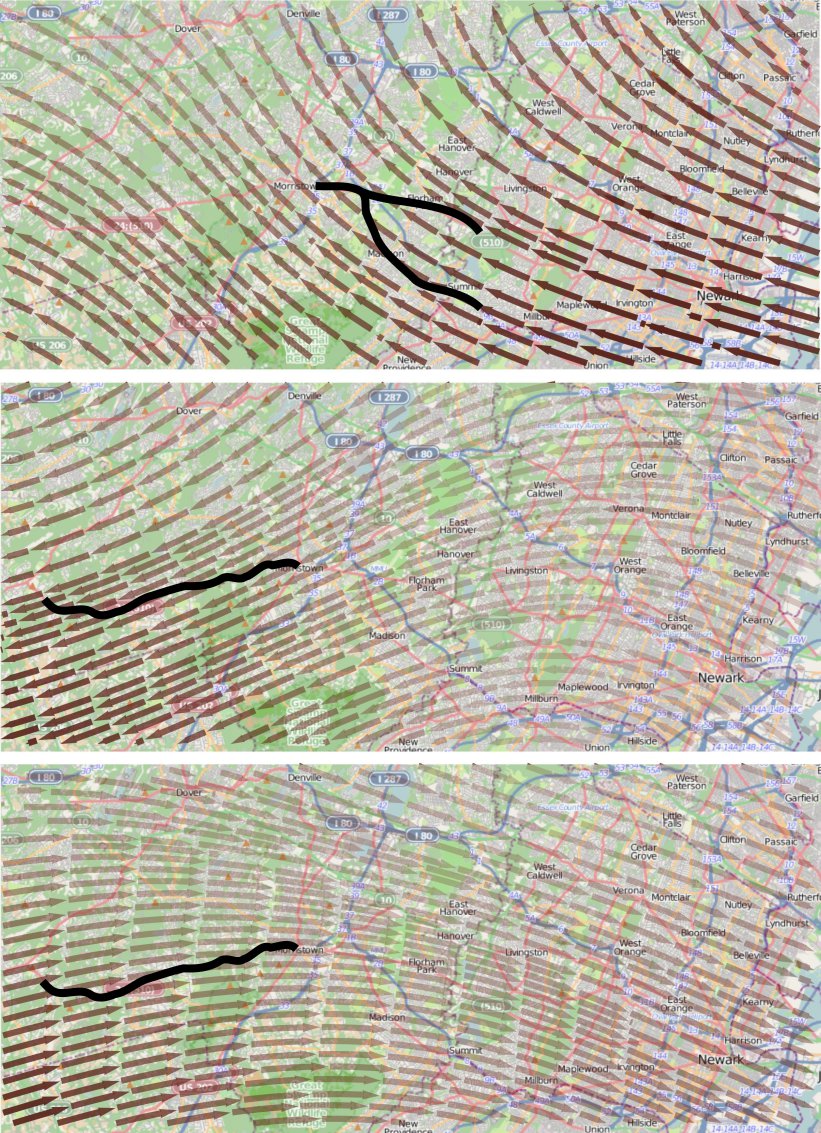}
\caption{\label{fig:cdr} Despite the noisy trajectories, \vfkm is able
to recover clear movement patterns related to highway (bold black lines) traffic around the city.
The three vector fields correspond to the clusters shown in Fig.~\ref{fig:handoffs}.}
\end{figure}

Given the sensitivity of CDR data, we took several steps to ensure the
privacy of individuals.
The data was collected and anonymized by a third party not involved in
the data analysis. 
Unlike other studies which replace phone numbers by anonymous unique
identifiers, we simply have no access to the
information~\cite{Becker:2011:RCU:2030112.2030130,becker2011tale}. In other
words, our dataset cannot associate multiple calls made by the
same individuals: each call is completely independent from all the
rest. The CDRs contain no information about the second party involved
in the call.
The only information available is the sequence of
antenna locations and handoff times for calls which were handled by
more than one physical antenna. Most calls are restricted to a single
antenna, so the dataset represents a small fraction of the total calls. 
In total, we collected over 370,000 calls over the period of a
single contiguous week of 2011. As we show in Figs.~\ref{fig:handoffs}
and~\ref{fig:cdr}, although the handoffs are quite noisy, \vfkm can
still recover movement patterns clearly related to the highway traffic
around the city.

\subsection{Comparison}
In this Section we compare our algorithm with the TraClus algorithm by 
Lee et al. \cite{lee2007trajectory} As discussed in Section~\ref{sec:related_words} the TraClus 
algorithm a density based algorithm that has been one of the main references
in the area of trajectory clustering. The algorithm consists in two main steps: trajectory simplification
in line segments and segment clustering. In the first step each given trajectory are simplified and
partitioned in line segments that approximate the input trajecty. The second step consists in clustering
the resulting line segments with a DBSCAN \nivan{is one reference need here?} like algorithm. 
The algorithm has essentially 2 parameters: a distance threshold called $\epsilon$ used to define
neighborhoods for each segment, and a density lower bound called $MinLns$ that is used to find neighborhoods
that define clusters, we refer the reader to the paper \cite{lee2007trajectory} for more details. One
important Traclus feature is that the time information is not used during the entire process and therefore
speed information is lost. 
For comparisons we use the C++ implementation of the algorithm that is available on the author webpage 
\footnote{\url{http://dm.kaist.ac.kr/jaegil/#Publications}}. In all the following experiments we use
the heuristics proposed on the paper (and also part of the author implementation) to select the parameter
values and when this heuristics fails to return a value we manually adjusted the parameters using the knowledge of the data set and visual inspection. We highlight that as, explained earlier, \vfkm and TraClus
mine different patterns and therefore it is difficult or maybe even impossible to say that one method
is better than the other all cases, but we do try to investigate what kind of patterns TraClus is not
able to mine that \vfkm and vice versa.

\begin{figure}[t]
\begin{center}
\subfigure[]{\label{figs:overview_trajectories}\includegraphics[width=.3\linewidth]{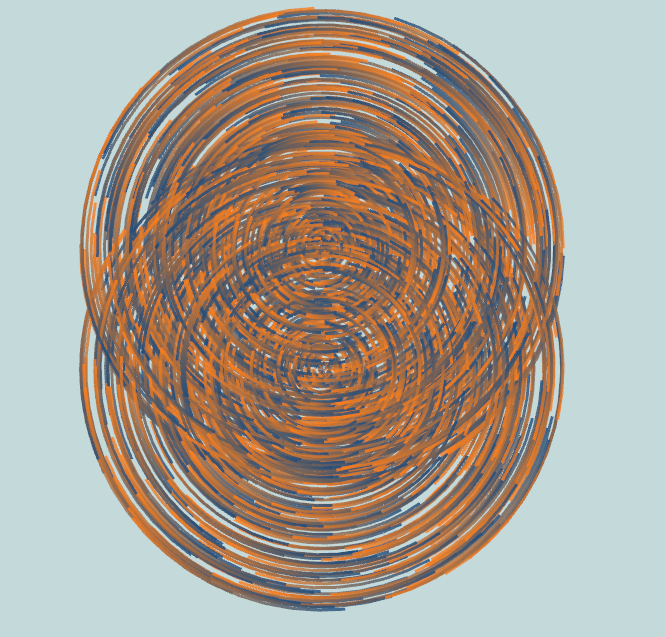}}
\subfigure[]{\label{figs:simplified_overview_trajectories}\includegraphics[width=.3\linewidth]{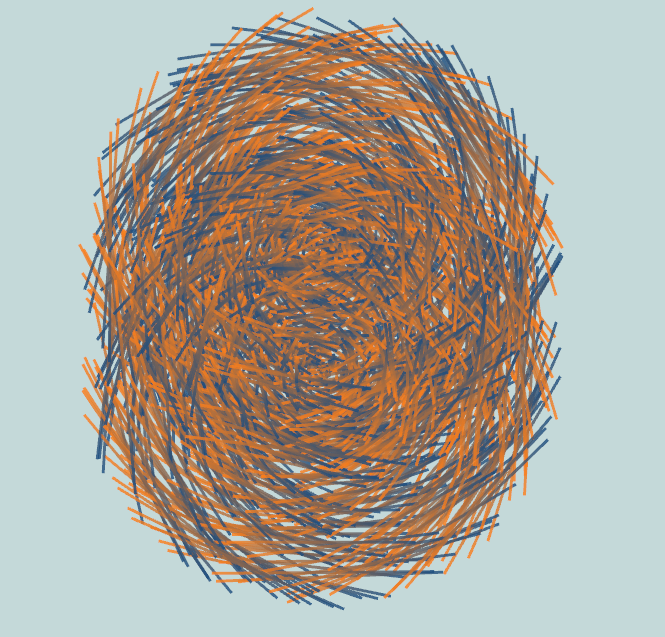}}
\subfigure[]{\label{figs:simplified_overview_trajectories_distance}\includegraphics[width=.3\linewidth]{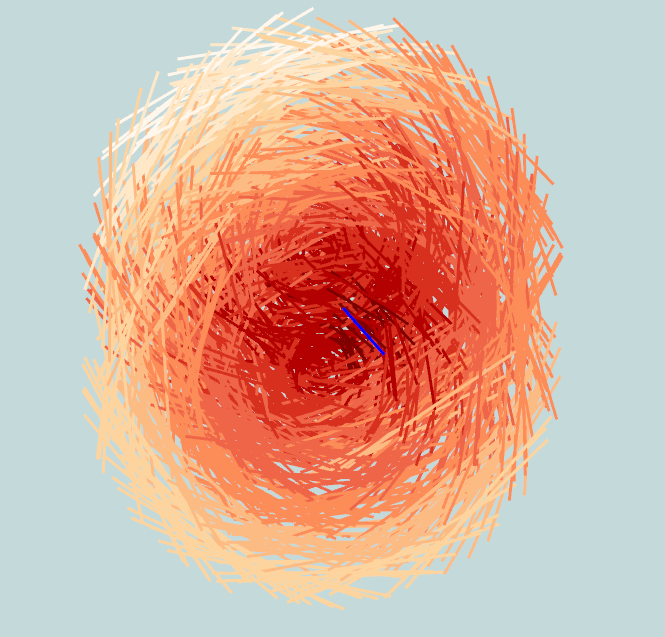}}
\end{center}
\caption{In (a) we show the input data set and in (b) we show the result of the first step of TraClus.
In this case, each trajectory results in exactly one segment (no partition happen). Figure (c) shows
the distance measure used by TraClus, the colors show the distances from the blue segment, where darker
colors mean smaller distances. We can see that both centers contain close segments to the blue segment
and therefore is impossible to completely separete the two centers using TraClus.}
\label{figs:traclus_synthetic_experiment}
\end{figure}

We first present the results of the TraClus algorithm on the synthetic dataset presented in 
Fig.~\ref{fig:overview}. In Fig.~\ref{figs:overview_trajectories} we show the data set used, and
in Fig.~\ref{fig:simplified_overview_trajectories} the simplified trajectories (result of the first 
step of TraClus). The idea of investigating this data set is to show that TraClus is not able to separate
the center since it looks at local features while as shown in Fig~\ref{fig:overview} \vfkm is able to 
capture the global structure by considering the vector fields. When the neighborhood size $\epsilon$ is 
set to be small, TraClus considers each segment as it's own cluster, making $\epsilon$ larger by the 
definition of the metric, the bottom part of the top center is grouped together with trajectories on the
bottom center. In this case we get the summarization problem highlighted in Fig.\ref{fig:summary_comparison},
in which the mean trajectory is not a good representative of the cluster.We present results of the experiments in Fig~\ref{figs:traclus_synthetic_experiment}. We can see that TraClus is unable to separete the two
centers.

\begin{figure}[t]
\begin{center}
\subfigure[]{\label{figs:overview_trajectories_268_clusters}\includegraphics[width=.25\linewidth]{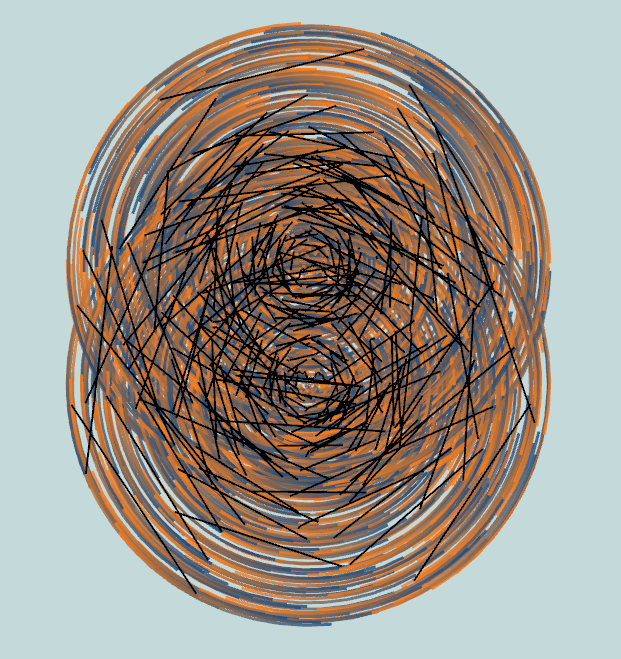}}
\subfigure[]{\label{figs:simplified_overview_trajectories_2_clusters}\includegraphics[width=.35\linewidth]{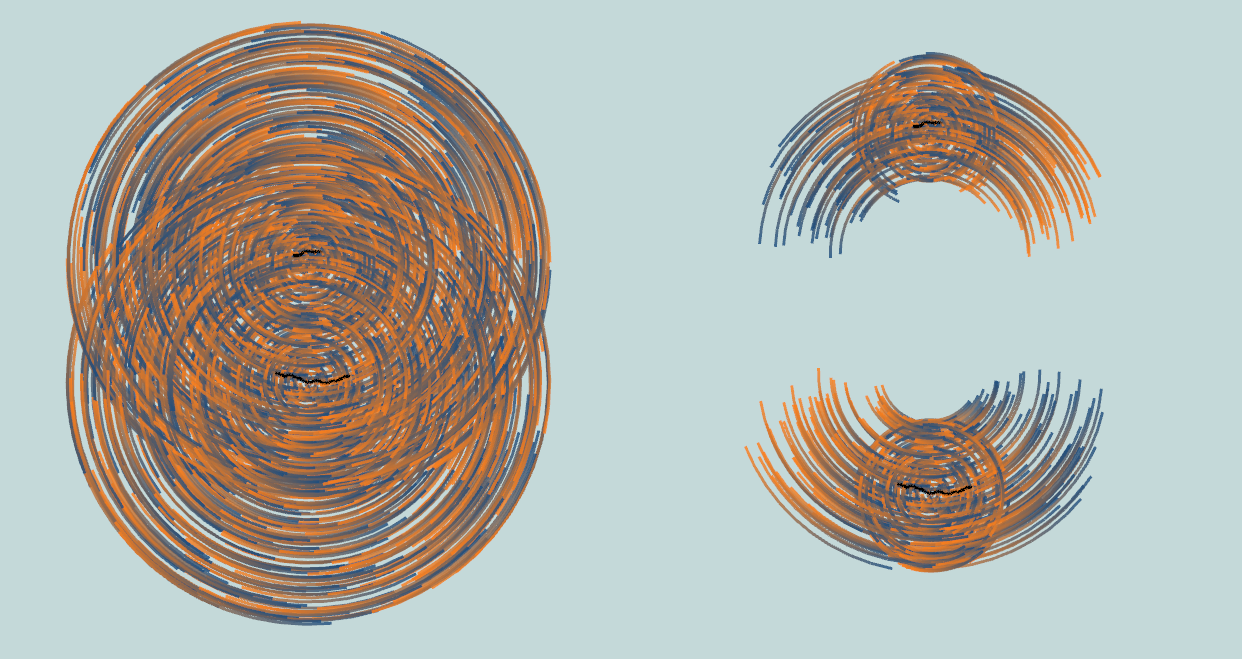}}
\subfigure[]{\label{figs:simplified_overview_trajectories_1_cluster}\includegraphics[width=.35\linewidth]{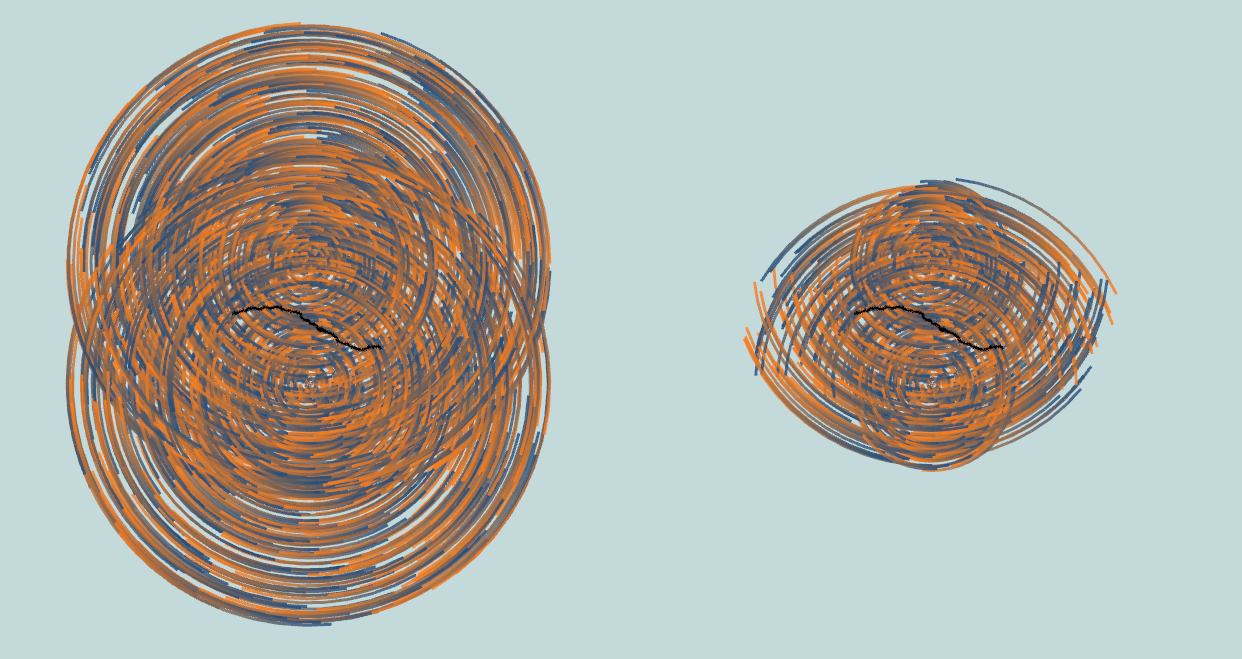}}
\end{center}
\caption{This figure illustrates the results of three experiments over the synthetic data set varying
the parameters. In (a) the parameters are $\epsilon = 0.03$ and $MinLns = 2$. In this case 268 are found.
Figure (b) shows the results for parameters $\epsilon = 0.23$ and $MinLns = 140$. In this experiment TraClus
is able to detect two cluster, but merges parts of the two centers, basically due to the distance issue
mentioned in Fig.~\ref{figs:simplified_overview_trajectories_distance}. A little variation on the parameters
($\epsilon = 0.25$ and $MinLns = 160$) causes TraClus to merge the two cluster in a single as shown in (c).
These results were obtained in 0.8, 1.6, and 6.5 seconds respectively.}
\label{figs:traclus_synthetic_experiment}
\end{figure}

\begin{figure}
\includegraphics[width=\linewidth]{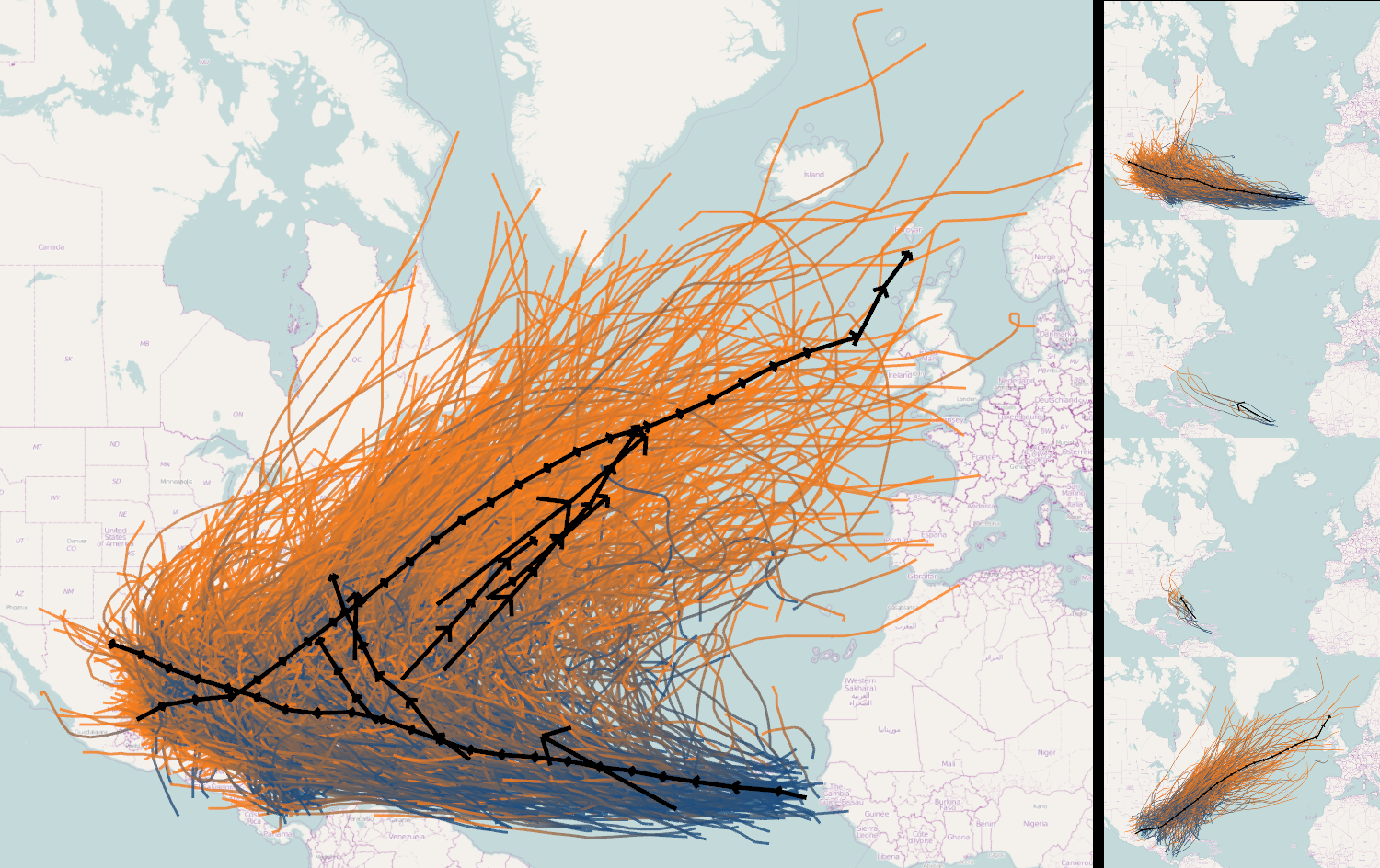}
\caption{On the left the nine clusters found. \nivan{need more comments}..}
\label{figs:traclus_hurricane_experiment}
\end{figure}

We now experiment with the hurricane data set discussed in Sec.~\ref{sec:experiments:hurricane}. 
Fig.~\ref{figs:traclus_hurricane_experiment} contains the results of TraClus on the hurricane 
dataset that are similar to the ones obtained by the authors \cite{lee2007trajectory}, notice
that our data set contains more trajectories this is the reason our results are slightly different.
By testing a range of parameters, We found that using the same parameters as the ones found by Lee
et al. gives us the best results. In this case TraClus detects 10 clusters, as shown on the left of
Fig.~\ref{figs:traclus_hurricane_experiment}. On the right of Fig.~\ref{figs:traclus_hurricane_experiment}
we show the details of some of the clusters. \nivan{missing R plots}. The results are obtained in 4 seconds.

\section{Discussion}\label{sec:discussion}

The algorithm proposed in this paper raises a number of interesting
questions, some of which we address in this section.
We will also discuss possible extensions of \vfkm in
context of its shortcomings or peculiarities.

\begin{figure}
\includegraphics[width=\linewidth]{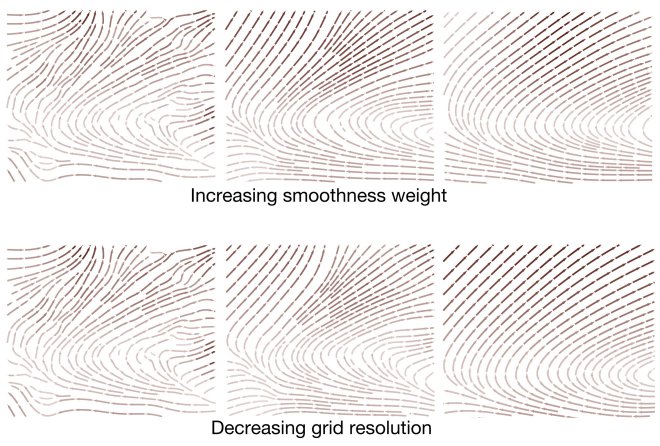}
\caption{\label{figs:smoothness}Increasing the smoothness weight in
  the optimization is essentially equivalent to decreasing the
  resolution. Top row, left to right: $R = 60$, $\lambda_L = [0.05, 0.95,
    0.9995]$. Bottom row, left to right: $\lambda_L = 0.05$, $R = 60, 20$, and $5$.}
\end{figure}

\subsection{Dependency on parameters}

In this section we briefly discuss our experience in how to
select the parameters in \vfkm.
We note, first of all, that although we can
select both grid resolution $R$ and the weight given to the Laplacian
regularization $\lambda_L$, in practice we never change $\lambda_L$, as
increasing $\lambda_L$ is basically the same as reducing $R$ (see Fig.~\ref{figs:smoothness}).
This happens for a well-known reason: the (orthogonal, unit-length)
eigenvectors of the Laplacian are naturally interpreted as
equivalent to the fundamental frequencies on the mesh, exactly like
sines and cosines are the fundamental frequencies on a circle~\cite{Vallet:2008:MH}. The
corresponding eigenvalues, on the other hand, are the (squares of) the
frequencies themselves. Because of this, as we increase $\lambda_L$ we
give larger weights to the eigenvectors corresponding to
high-frequency signals, and the system tends towards lower-frequency
results. At the same time, reducing $R$ 
directly band-limits the signal on the vector field, which is
a quite similar effect. As a result, we set our $\lambda_L$ to
be $0.05$ in all of our experiments and vary $R$ instead. This has the
distinct advantage of generating much smaller linear systems, which can
be solved much more quickly.

Picking an appropriate number of clusters remains generally an open
problem even in the case of traditional $k$-means, and we offer no
substantive contributions on that matter. Many methods proposed in the
literature try to attack this problem (for example see
\cite{Fang:2012:SNC:2064097.2064130} and references therein), however no definitive
algorithm solves this problem optimally for all aplications in general
settings.  Still, we stress that as far as performance is concerned,
\vfkm compares quite favorably to results reported in the
literature. It is much easier to include a human analyst in the loop
and make cluster count an interactive procedure with \vfkm than with
other methods.

\subsection{Advantages}

Our proposed model strikes a nice balance between richness and
expressivity of features, and simplicity of implementation and
analysis. We believe this is a significant advantage over the current
methods for trajectory clustering. As mentioned earlier, by representing
the cluster centers as vector fields and using those as a means to define
similary between trajectories, we can eliminate expensive computations
of metrics for trajectories and the computations of the centroid trajectory
as well \cite{gudmundsson2012computational}. \VFKM is also potentially highly
parallelizable. Our prototype includes no significant optimizations,
but it is obvious that separate components of vector fields can
be computed in parallel, and that many of the intermediate matrices in
the linear solvers can be reused from one iteration to the next. We
expect these to further increase the performance, and allow \vfkm to
handle even larger datasets.

\subsection{Limitations}

Since \vfkm is akin to \km, it inherits the good and bad features from
it as well. Still, \km is a very well-studied algorithm and many
solutions developed to avoid problems in \km can be adapted to our
work with \vfkm.  For example, the choice of the initial clusters have
a big impact in the results achieved by \vfkm.  Many techniques have
been proposed to choose good initial centers for the clusters.  We
highlight the $k$-means++ approach proposed by Arthur et
al.~\cite{arthur2007k}, which consists of defining the
initial centers by randomly choosing the points with probability
proportional to the square of the distance between this point and the
closest centers already defined. The initialization step used in our
implementation of \vfkm (Section~\ref{sec:initialization}) has no
theoretical guarantees, while $k$-means++ gives a $\log k$
approximation to the $k$-means clustering problem. It is an
interesting avenue of future work to investigate if the statements
about $k$-means++ carry over to \vfkm.
Other initialization approaches have also been proposed by Yedla et
al. and Patel and
Mehta~\cite{yedla2010enhancing,patel2011hierarchical}. 
He et al. offer another general study on
possible approaches for cluster initialization~\cite{he2004initialization}.

Another issue that happens in \km and also may arise during the
execution of the \vfkm is the singularity problem
\cite{pakhira2009modified}. This occurs when one or more clusters
become empty during the computation.  This problem is due to bad
initialization that may arise in the usual \km and can arise in \vfkm
as well.  Our current implementation simply repopulates the empty
cluster by splitting the largest cluster at that point in the
optimization. This is entirely \emph{ad-hoc}, and we would like a
better solution. Again, one could also adapt the methods proposed to avoid
this problem for \km to work in \vfkm.  Pahkira has some proposal
to avoid empty clusters, and points to further
references~\cite{pakhira2009modified}.

\section{Extensions and Future Work}


The version of \vfkm presented in this paper derives steady vector
fields from trajectory data only.  However, it can be generalized to
produce time varying vector fields. Fundamentally, this imposes no
problems: one simply creates a three-dimensional grid and sets the
constraints on the interior of tetrahedral decompositions of a regular
grid. It remains to be seen, however, whether the performance
characteristics that are so attractive about our current algorithm
will remain so in a three-dimensional extension.

For the sake of simplicity, we described the \vfkm algorithm in Section~\ref{sec:method} in
two dimensions. Nevertheless, we note that \vfkm works with trajectories in any dimension $d$.  More precisely, one still could
use a variant of the formulation in Equation~\ref{eq:formulation}. 
One would need to involve the region on interest in a 
simplicial grid of dimension $d$, in which we assume linear interpolation
inside each simplex of the grid.  
Most of the steps would be straightforward to carry through. We note,
however, that to the best of our knowledge there is no good
three-dimensional equivalent to the cotangent weight of the
two-dimensional Laplacian.

Many real-life applications require alignment not only of tangent
directions and speed, but also of occurrences in time. Our current
algorithm cannot handle these constraints. However, we believe this
can be addressed by including a \emph{time scalar field} as an
additional field to be constructed from constraints. From there, the
assignment step would simply consider time mismatches as another
penalty.

Another interesting research direction worthy of exploring in future
work is to apply \vfkm using user input. One could imagine that the
user could visually define the vector field or even some sample
trajectories and the algorithm would retrieve all trajectories in the
database that follow those patterns given by the user, \ie the
ones for which the error relative to the vector field obtained from
the trajectories given by the user is small enough.

\section*{Acknowledgments}
We would like to thank the OpenStreetMap project \cite{haklay2008openstreetmap}
for providing the maps used in many of the figures in this paper. 
 This work was supported in part by the National Science Foundation (CCF-08560, CCF-
0702817, CNS-0751152, CNS-1153503, IIS-0844572, IIS-
0904631, IIS-0906379, IIS-1153728, and NIH ITKv4), the
Department of Energy and IBM Faculty Awards. 

\bibliographystyle{plain}
\bibliography{paper}


\end{document}